# Employing Explainable Artificial Intelligence (XAI) Methodologies to Analyze the Correlation between Input Variables and Tensile Strength in Additively Manufactured Samples

Akshansh Mishra[1], Vijaykumar S Jatti[2]

[1]School of Industrial and Information Engineering, Politecnico Di Milano, Milan, Italy

[2] Symbiosis Institute of Technology, Symbiosis International (Deemed) University, Pune, Maharashtra, 412115, India

**Abstract:** This research paper explores the impact of various input parameters, including Infill percentage, Layer Height, Extrusion Temperature, and Print Speed, on the resulting Tensile Strength in objects produced through additive manufacturing. The main objective of this study is to enhance our understanding of the correlation between the input parameters and Tensile Strength, as well as to identify the key factors influencing the performance of the additive manufacturing process. To achieve this objective, we introduced the utilization of Explainable Artificial Intelligence (XAI) techniques for the first time, which allowed us to analyze the data and gain valuable insights into the system's behavior. Specifically, we employed SHAP (SHapley Additive exPlanations), a widely adopted framework for interpreting machine learning model predictions, to provide explanations for the behavior of a machine learning model trained on the data. Our findings reveal that the Infill percentage and Extrusion Temperature have the most significant influence on Tensile Strength, while the impact of Layer Height and Print Speed is relatively minor. Furthermore, we discovered that the relationship between the input parameters and Tensile Strength is highly intricate and nonlinear, making it difficult to accurately describe using simple linear models.

1. ## Introduction

Additive Manufacturing (AM), commonly known as 3D printing, is a rapidly expanding technology that offers efficient and cost-effective production of intricate designs, surpassing traditional manufacturing methods. However, the success of AM heavily relies on optimizing process parameters and identifying optimal designs. This is where Artificial Intelligence (AI) revolutionizes the AM industry. One crucial factor in AM is the selection and optimization of process parameters like temperature, speed, and layer height. AI aids in identifying optimal process parameters by analyzing data from diverse sources such as machine sensors, simulation software, and historical data. Machine learning algorithms detect patterns and correlations between process parameters and product quality, optimizing efficiency and cost-effectiveness. Quality control is another vital aspect of AM, which can be challenging due to design complexity and parameter variability. AI assists in identifying defects and anomalies by analyzing sensor and camera data, facilitating real-time monitoring, feedback, and proactive defect correction.

AM offers unparalleled design freedom, enabling the production of complex shapes unattainable through traditional means. However, designing for AM requires expertise in the technology and an understanding of design constraints and possibilities. AI aids in design optimization by generating and evaluating designs using generative algorithms and optimization techniques. This streamlines the process, identifying designs that meet specifications while minimizing production time and costs.

Researchers have proposed innovative approaches using machine learning in AM applications. For example, Kononenko et al. [15] proposed the utilization of acoustic emission (AE) signals and machine learning techniques for in situ fracture identification of AM-fabricated parts. By distinguishing crack AE events from background noise, they built classification ML models achieving up to 99% accuracy. Mukherjee et al. [16] presented a novel method to obtain subsurface temperature distribution metrics during laser melting, utilizing in situ synchrotron X-ray diffraction observations and supervised machine learning surrogate models. Hemmasian et al. [17] used Flow-3D simulation software to create datasets and trained a convolutional neural network to predict the three-dimensional temperature field based on process parameters and time step inputs.

As AI expands, transparency and interpretability become vital. Explainable Artificial Intelligence (XAI) addresses this concern by providing insights into AI algorithms' inner workings and decision-making processes. XAI has broad applications across various industries, including additive manufacturing. By implementing XAI, manufacturers gain visibility into the factors influencing the quality of finished products, such as temperature, humidity, and printing speed. Analyzing this data enables process optimization for consistent quality and waste reduction. XAI also evaluates finished product quality, detecting defects and errors in the printing process in real-time. With prompt feedback on product quality, manufacturers ensure adherence to required standards. XAI can analyze data from 3D printers, identifying patterns indicative of impending equipment failure. By detecting potential issues early, manufacturers can conduct maintenance and repairs, minimizing downtime and enhancing productivity.

This study investigates the influence of specific input parameters, namely Infill percentage, Layer Height, Extrusion Temperature, and Print Speed, on the resulting Tensile Strength in additively manufactured specimens. The primary aim of this research is to gain a deeper understanding of the relationship between these input parameters and Tensile Strength, and to identify the key parameters that significantly affect the performance of the additive manufacturing process.

## 2. Experimental Procedure

In this research, the Fused Deposition Modeling (FDM) samples were manufactured using a Creality Ender 3 printer with a build area measuring 220 × 220 × 250 mm³. The parts were designed using CATIA software and then converted into the STL format before being sliced into machine-readable g-code using the Cura engine of the Repetier software. The tensile samples were created in accordance with the ASTM D638 standard, with dimensions of 63.5 × 9.53 × 3.2 mm³. For the production of these samples, polylactic acid (PLA) was chosen as the material, as it is commonly used for FDM-printed parts. The strength of the samples was determined through uniaxial tensile tests, and the dimensions of the tensile samples are illustrated in Figure 1.

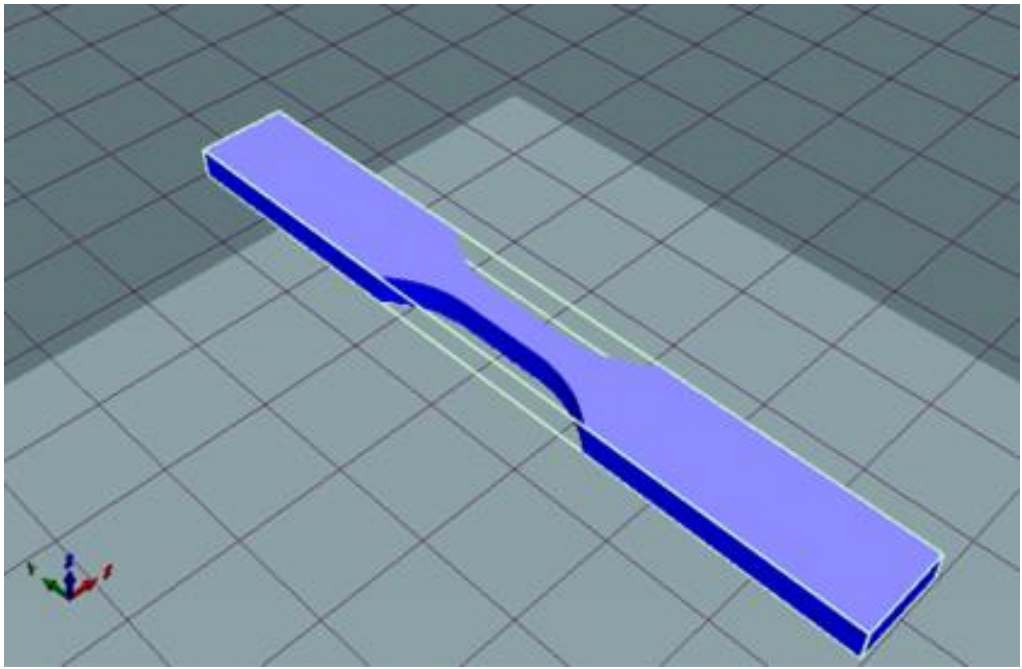

Figure 1: Dimensional sketch of Tensile Specimen

The FDM Test specimen printing setup is shown in Figure 2.

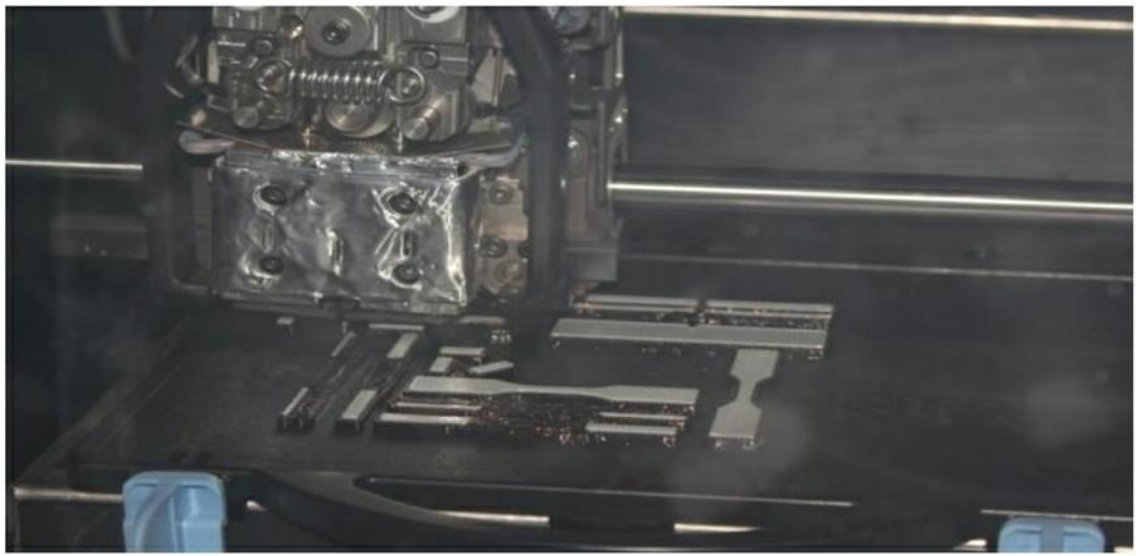

Figure 2: FDM Test specimen printing setup

This study focuses on examining four primary input parameters: Infill percentage, Layer height, Print speed, and Extrusion temperature. Infill percentage refers to the amount of material inside a fabricated part, determining its density based on specific requirements. The study incorporates infill percentages of 10%, 33%, 55%, 78%, and 100%. Layer height in Fused Deposition Modeling (FDM) corresponds to the thickness of a single deposited layer and utilizes a nozzle diameter of 0.4 mm. The layer heights tested are 0.08 mm, 0.16 mm, 0.24 mm, 0.32 mm, and 0.4 mm. Print speed represents the rate at which material is deposited by the nozzle and can impact physical part wear or material distribution. The research examines print speeds of 20 mm/s, 35 mm/s, 50 mm/s, 65 mm/s, and 80 mm/s. Extrusion temperature, controlled by the extruder's heating system, affects material viscosity and is explored at temperatures of 190 °C, 200 °C, 210 °C, 220 °C, and 230 °C. The study maintains the following constant parameters: Infill pattern (Honeycomb), Bed temperature (80 °C), Build orientation (X-0 degree, Y-0 degree), Raster angle (0/90 degree), shell thickness (0), raster width (0.44), and air gap (0).

Following mechanical testing, the experimental dataset is collected and saved as a CSV file. This file is then imported into the Google Colaboratory environment for the implementation of the Python-based Explainable Artificial Intelligence (XAI) framework developed. Figure 3 illustrates the framework employed in this study.

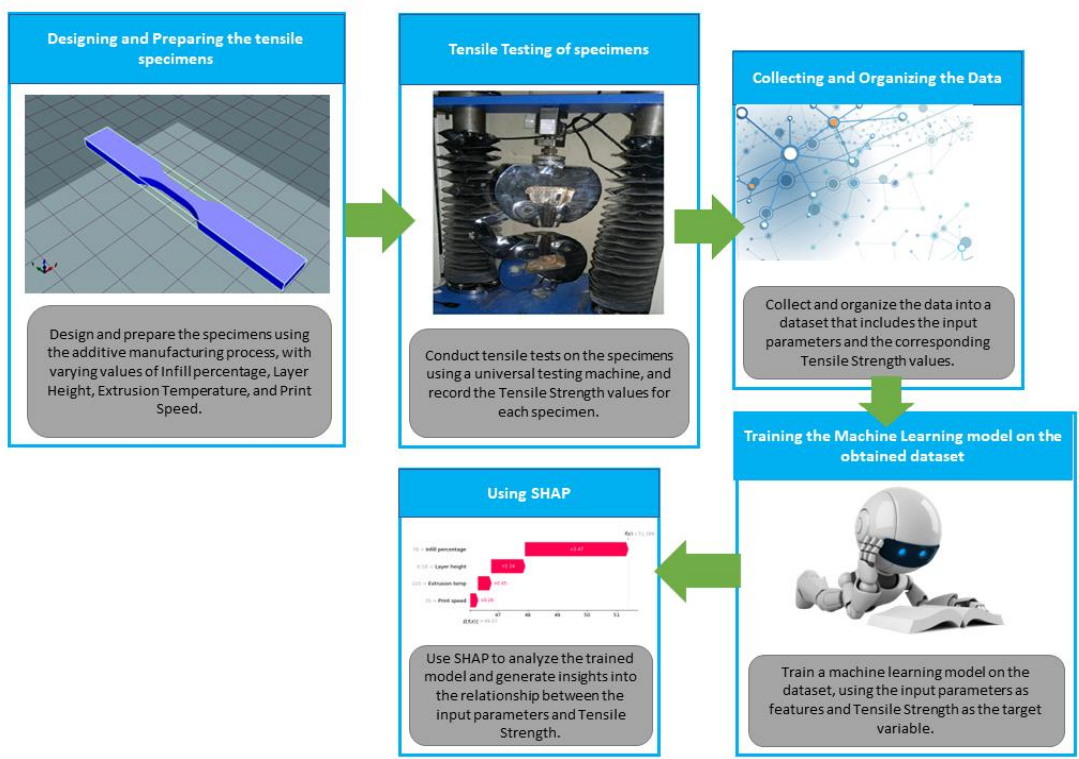

Figure 3: XAI Framework implemented in the present study.

Analyzing and interpreting the results of a machine learning model is an essential aspect of model evaluation. One widely used approach for interpreting machine learning models is SHAP (SHapley Additive exPlanations). SHAP is a method that offers both local and global explanations for understanding individual predictions and overall model behavior. In the following section, we will delve into the mechanics of SHAP and how it identifies the input parameters that impact the output parameter of a machine learning model.

### 3. Results and Discussion

The experimental results obtained are displayed in Table 1, indicating the input parameters as Infill percentage, Layer height, print speed, and Extrusion Temperature, while the output parameter is Tensile Strength.

Table 1: Experimental Results.

| Infill percentage | Layer height (mm) | Print speed (mm/s) | Extrusion temperature ( ℃ ) | Tensile strength (MPa) |
|---|---|---|---|---|
| 78 | 0.32 | 35 | 220 | 46.17 |
| 10.5 | 0.24 | 50 | 210 | 42.78 |
| 33 | 0.16 | 35 | 220 | 45.87 |
| 33 | 0.32 | 35 | 200 | 41.18 |
| 33 | 0.16 | 65 | 200 | 43.59 |
| 100 | 0.24 | 50 | 210 | 54.2 |
| 78 | 0.16 | 35 | 200 | 51.88 |
| 33 | 0.32 | 65 | 200 | 43.19 |
| 78 | 0.32 | 65 | 200 | 50.34 |
| 33 | 0.16 | 65 | 220 | 45.72 |
| 78 | 0.16 | 35 | 220 | 53.35 |
| 55.5 | 0.24 | 50 | 210 | 49.67 |
| 33 | 0.32 | 35 | 220 | 45.08 |
| 55.5 | 0.24 | 50 | 190 | 47.56 |
| 55.5 | 0.24 | 50 | 210 | 48.39 |
| 78 | 0.32 | 65 | 220 | 46.49 |
| 55.5 | 0.24 | 50 | 210 | 47.21 |
| 55.5 | 0.24 | 50 | 210 | 48.3 |
| 55.5 | 0.24 | 50 | 230 | 50.15 |
| 33 | 0.32 | 65 | 220 | 43.35 |
| 55.5 | 0.24 | 50 | 210 | 45.33 |
| 55.5 | 0.24 | 80 | 210 | 45.56 |
| 78 | 0.16 | 65 | 200 | 49.84 |
| 55.5 | 0.24 | 20 | 210 | 48.51 |
| 55.5 | 0.08 | 50 | 210 | 42.63 |
| 55.5 | 0.4 | 50 | 210 | 42.87 |
| 55.5 | 0.24 | 50 | 210 | 47.14 |
| 78 | 0.32 | 35 | 200 | 45.17 |
| 55.5 | 0.24 | 50 | 210 | 47.07 |
| 78 | 0.16 | 65 | 220 | 50.99 |
| 33 | 0.16 | 35 | 200 | 51.55 |

The impact of the "Infill percentage" input parameter on the predictions of a trained machine learning model is examined using the partial dependence plot (PDP) method from the SHAP library. The PDP allows for an investigation of the relationship between a specific input feature, in this case, the "Infill percentage," and the output of the machine learning model while keeping all other input features constant at a particular value.

The findings derived from the PDP plot, illustrated in Figure 4, reveal the association between the "Infill percentage" input parameter and the output of the machine learning model. The PDP plot showcases a line graph, where the x-axis represents the infill percentage, and the y-axis displays the expected value of the model output. The line depicted in the plot represents the average effect of the "Infill percentage" input feature on the model output, with all other input features held at a constant level. From a research perspective, the PDP plot offers valuable insights into the significance of the "Infill percentage" input feature in predicting the machine learning model's output. It also aids in the identification of any nonlinear relationships between the input feature and the output, which may not be discernible through simple scatter plots or correlation coefficients. Researchers can leverage the PDP plot to optimize the "Infill percentage" input feature to achieve the desired output from the machine learning model.

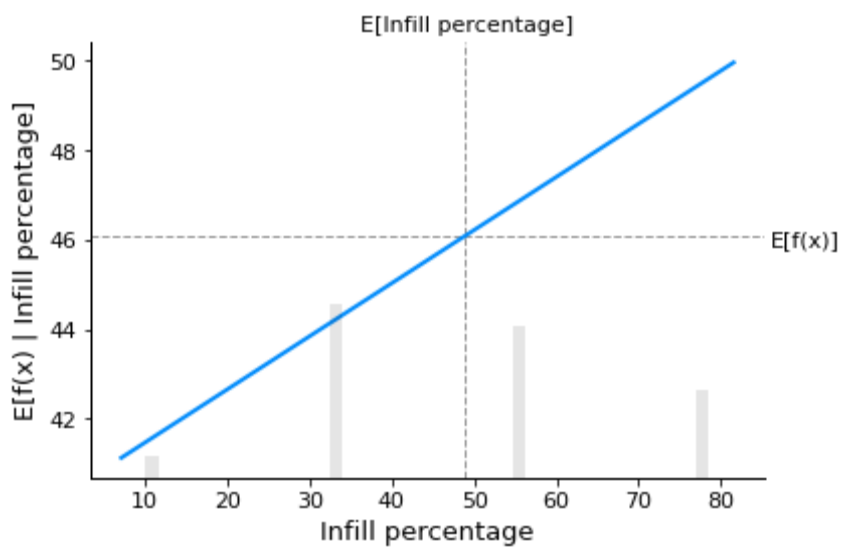

Figure 4: Obtained Partial Dependence plot

To calculate SHAP values for a given machine learning model, the shap.Explainer function is utilized. This function requires two arguments: the model.predict function, which generates predictions from the model, and the X20 dataset. Using these arguments, an explainer object is created to compute the SHAP values for the entire dataset X using the explainer(X) function. The resulting shap_values represent the SHAP values for each feature in every sample within the dataset.

To visualize the relationship between the "Infill percentage" feature and the model's output, the shap.partial_dependence_plot function is employed. This function incorporates the computed SHAP values, accounting for their impact. The resulting plot is a partial dependence plot (PDP) that illustrates the influence of the "Infill percentage" feature on the model's predictions, as depicted in Figure 5.

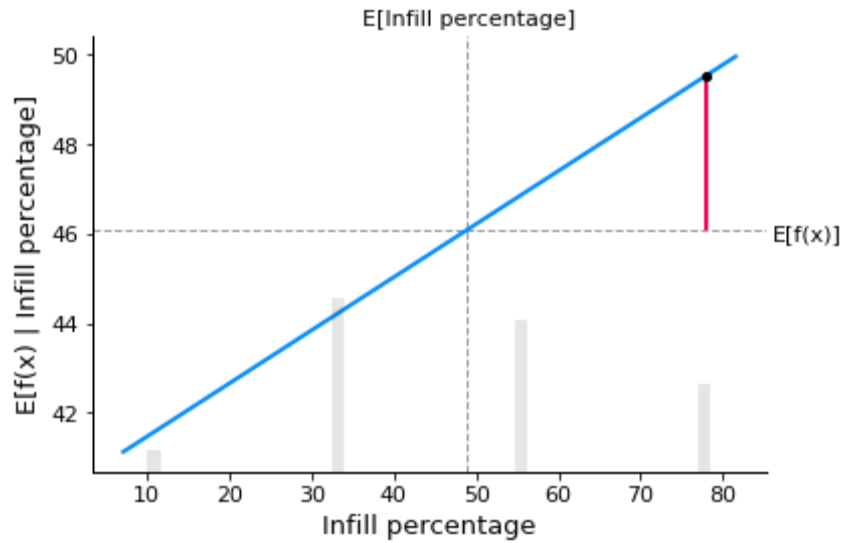

Figure 5: Plot obtained while taking into the account of SHAP Values

The result interpreted from the PDP plot shows the relationship between the "Infill percentage" input feature and the output of the machine learning model while accounting for the SHAP values. This means that the PDP plot displays the average effect of the "Infill percentage" input feature on the model output, while taking into account the interactions between the "Infill percentage" feature and the other input features.

Compared to the previous plot i.e. in Figure 4, this interpretation uses the SHAP values to compute the PDP plot, while the previous interpretation used the average effect of the feature across all samples in the dataset. This difference can lead to more accurate and informative PDP plots, especially when dealing with complex machine learning models and datasets. Additionally, this interpretation allows for the examination of the interaction effects between the "Infill percentage" feature and the other input features, which may not be possible with other methods.

The waterfall plot shown in Figure 6 displays each feature's contribution as a horizontal bar, with positive contributions shown in blue and negative contributions shown in red. The plot also includes a vertical line representing the baseline value (i.e., the base_values), and the final prediction is shown at the top of the plot. By analyzing the plot, we can interpret how each feature's value affects the model's prediction for the specific sample. Features with positive (blue) contributions increase the prediction, while features with negative (red) contributions decrease it. The sum of all feature contributions, plus the baseline value, should equal the final prediction shown in the plot.

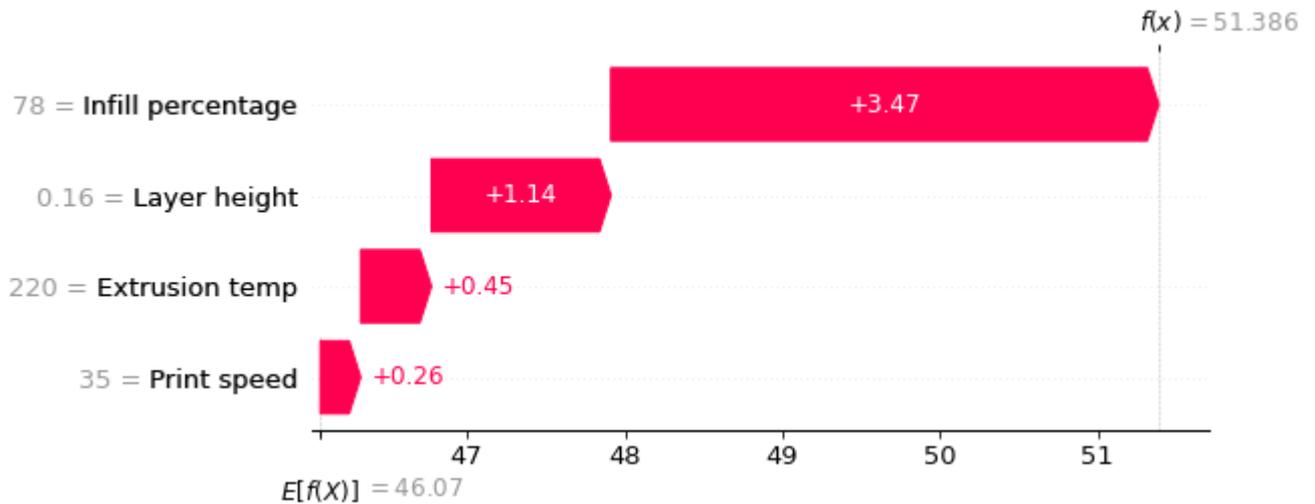

Figure 6: Obtained Waterfall plot

A Generalized Additive Model (GAM) is a type of machine learning model commonly used for regression analysis. Unlike linear regression models, GAMs allow for more complex relationships between input variables and output variable by adding non-linear functions of input variables. In SHAP, a GAM model is utilized to provide explanations for complex and opaque machine learning models. The SHAP framework is a well-known method used to interpret the predictions of machine learning models. It simplifies the opaque model by approximating it with a simpler model that is easier to understand.

The SHAP library includes an implementation of a GAM model that can be used to generate explanations for a wide range of machine learning models. The GAM model is trained on a portion of the training data to approximate the behavior of the opaque model. SHAP values are then computed based on the differences between the predictions of the GAM model and the actual predictions of the opaque model. The resulting SHAP values offer insights into the most significant features that contribute to the predictions of the opaque model, and how these features influence the final prediction. This information can be used to understand the behavior of the model, uncover potential biases or errors, and improve the overall accuracy and reliability of the machine learning system.

The partial dependence plot depicted in Figure 7 shows the average effect of changing the "Infill percentage" feature on the model's predicted output, while holding all other features at their mean values. The SHAP value overlay on the plot indicates how much of the observed effect of changing the "Infill percentage" feature is due to the "Infill percentage" feature itself (i.e., the direct effect), and how much is due to interactions with other features (i.e., the indirect effects).

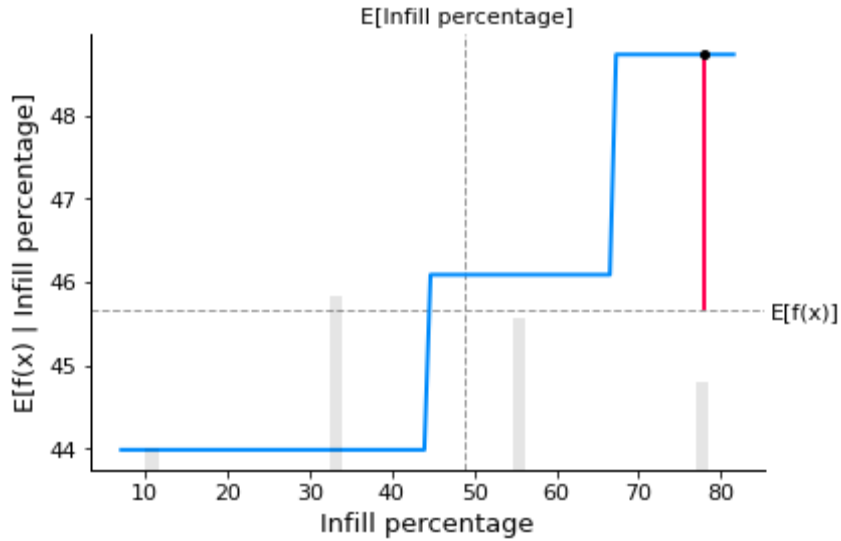

Figure 7: Partial Dependence plot obtained after implementation of GAM model

The updated waterfall plot is displayed in Figure 8. The introduction of a GAM (Generalized Additive Model) model in this particular case can potentially alter the importance of input features. GAM models have the capability to capture intricate, nonlinear relationships between input features and the output variable. Consequently, they can provide a more accurate depiction of the true relationship between input features and the output variable compared to simpler models like linear regression.

In the context of the SHAP library, the GAM model is employed to generate explanations for predictions made by a black-box machine learning model. The GAM model is trained on a subset of the training data and approximates the behavior of the black-box model. SHAP values are then computed based on the disparities between the predictions of the GAM model and the actual predictions of the black-box model.

The resulting SHAP values offer insights into which features hold the most significance for the predictions of the black-box model and how they contribute to the final prediction. The variation in input feature importance after implementing the GAM model may arise because the GAM model assigns different levels of importance to certain input features compared to other models. This divergence could be attributed to the GAM model's ability to capture complex relationships between input features and the output variable that other models may not capture. As a result, the changes in input feature importance observed after implementing the GAM model in this case can be attributed to the model's capacity to grasp intricate relationships between the input features and the output variable, ultimately leading to a more accurate representation of their true relationship.

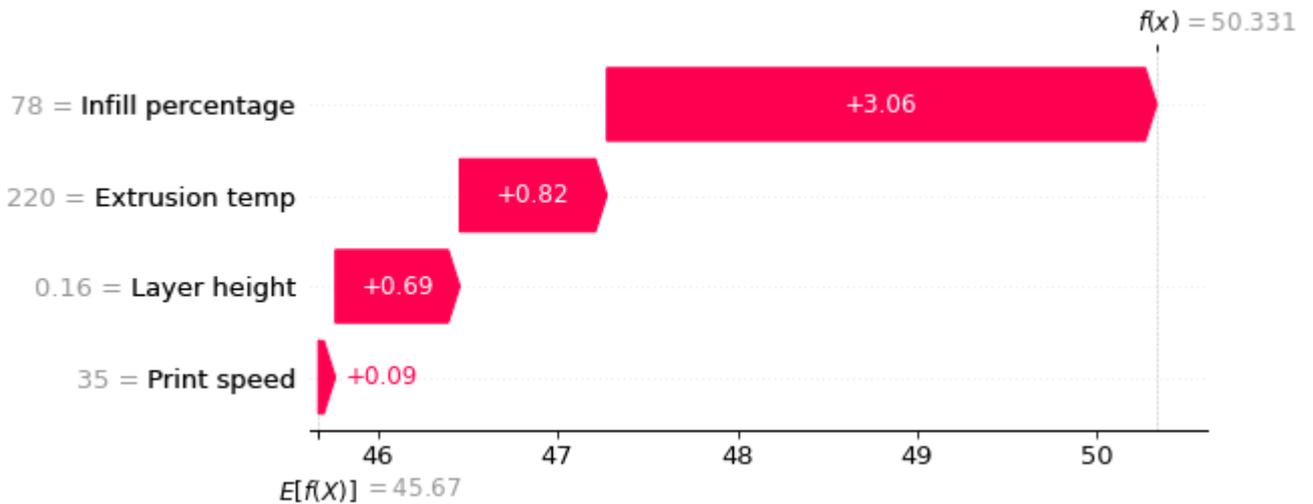

Figure 8: Updated waterfall plot after the implementation of GAM model

The SHAP values for all input features of a machine learning model, explained using a GAM model in SHAP, are visualized in Figure 9 through a beeswarm plot. These SHAP values indicate the contribution of each input feature to the model's prediction for a specific sample. The beeswarm plot illustrates the distribution of SHAP values for each feature, where the x-axis represents the SHAP value and the y-axis corresponds to the respective feature.

A positive SHAP value suggests that the feature contributes to a higher prediction value, while a negative value implies that the feature contributes to a lower prediction value. The magnitude of the SHAP value indicates the feature's importance for the model's prediction. A larger absolute SHAP value indicates a higher importance. To generate the plot, the shap.plots.beeswarm function is applied to the shap_values_ebm object, which stores the SHAP values computed using the GAM model. The resulting beeswarm plot provides an overview of the SHAP value distributions for all input features in the model.

This beeswarm plot is valuable for identifying the most influential features in the model's predictions and understanding their respective contributions. It can help detect potential biases or errors in the model and guide feature selection and engineering to enhance the model's performance.

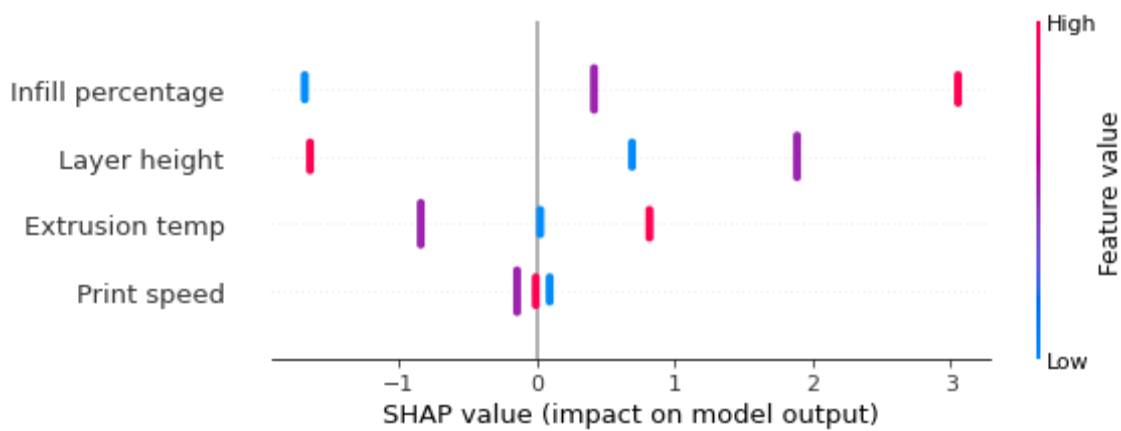

Figure 9: Obtained Beeswarm plot

The heatmap displayed in Figure 10 portrays the SHAP values, with the color of each cell indicating the magnitude of the SHAP value. Red hues represent positive values, while blue hues represent negative values. The intensity of the color corresponds to the absolute value of the SHAP value, indicating the relative importance of the feature for the model's prediction on that specific sample.

Visualizing the SHAP values in a heatmap allows for the identification of patterns and trends within the data. It facilitates the observation of how different input features contribute to the model's predictions across various samples. The heatmap enables the detection of potential biases or errors in the model and facilitates exploration of the relationships between the input features and the output variable. By examining the heatmap, insights can be gained into which features strongly influence the model's predictions and how they interact with each other, providing a comprehensive understanding of the model's behavior.

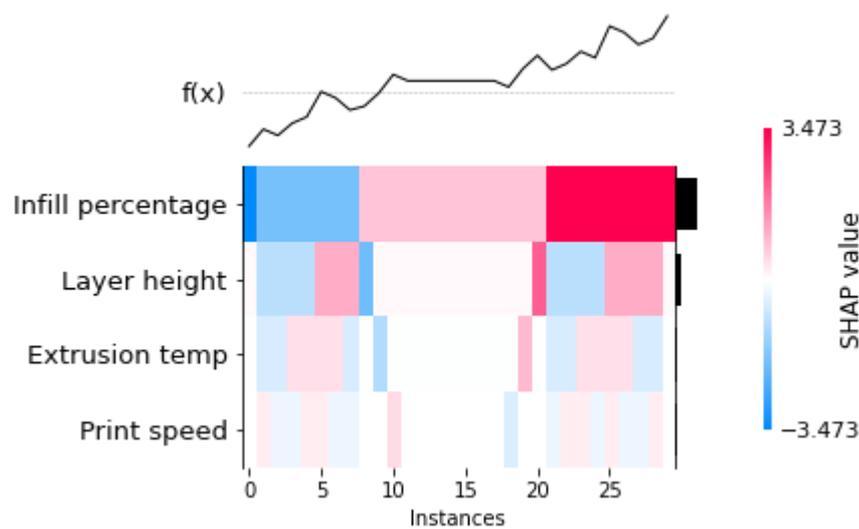

Figure 10: Obtained Heat Map

By examining the heatmap, we gain insights into the key input features that hold the highest significance for the model's predictions and understand their respective contributions to the final prediction. Additionally, we can uncover patterns or trends within the data, such as correlations between different input features. This analysis enables the detection of potential model-related concerns, such as overfitting or underfitting, which may impact the model's performance and reliability. The heatmap serves as a valuable tool for assessing the model's behavior, identifying influential features, and unveiling relationships and potential issues that require attention in order to improve the model's accuracy and robustness.

4. Conclusion

In conclusion, this study utilized Explainable Artificial Intelligence techniques to examine the impact of input parameters on the Tensile Strength of additively manufactured specimens. By employing the SHAP framework, valuable insights were obtained regarding the system's behavior. The findings highlighted the significant influence of Infill percentage and Extrusion

Temperature on Tensile Strength, whereas Layer Height and Print Speed had relatively minor effects. Furthermore, the study discovered that the relationship between input parameters and Tensile Strength was intricate and non-linear, surpassing the capabilities of simple linear models.

Moving forward, there are several avenues for future exploration in this field. One area of focus is optimizing the input parameters to enhance the Tensile Strength of additively manufactured specimens. Further research could aim to identify the optimal combinations of input parameters to achieve the desired Tensile Strength. Additionally, extending the study to other output parameters of additive manufacturing processes, such as surface finish, dimensional accuracy, and mechanical properties, would deepen the understanding of process behavior.

Moreover, there is scope to explore more advanced machine learning models that can accurately predict the system's behavior and improve the overall performance of additive manufacturing processes. By delving into these areas, researchers can advance the field of additive manufacturing and pave the way for enhanced process optimization and better-quality output.

**Conflict of Interest**

Authors decalre that they don't have any conflict of interest.